\documentclass[conference]{IEEEtran}
\IEEEoverridecommandlockouts
\pdfoutput=1
\usepackage{cite}
\usepackage{amsmath,amssymb,amsfonts}
\usepackage{algorithmic}
\usepackage{graphicx}
\usepackage{textcomp}
\usepackage{xcolor}
\usepackage{subcaption}
\usepackage{algorithm}
\usepackage{multirow}
\usepackage{slashbox}

\def\BibTeX{{\rm B\kern-.05em{\sc i\kern-.025em b}\kern-.08em
    T\kern-.1667em\lower.7ex\hbox{E}\kern-.125emX}}
\begin{document}

\title{Modular Federated Learning
}

\author{\IEEEauthorblockN{Kuo-Yun Liang}
\IEEEauthorblockA{\textit{Connected Systems} \\
\textit{Scania CV AB}\\
S\"odert\"alje, Sweden\\
kuo-yun.liang@scania.com}
\and
\IEEEauthorblockN{Abhishek Srinivasan}
\IEEEauthorblockA{\textit{Connected Systems} \\
\textit{Scania CV AB}\\
S\"odert\"alje, Sweden\\
abhishek.srinivasan@scania.com}
\and
\IEEEauthorblockN{Juan~Carlos Andresen}
\IEEEauthorblockA{\textit{Connected Systems} \\
\textit{Scania CV AB}\\
S\"odert\"alje, Sweden\\
juan-carlos.andresen@scania.com}
}

\maketitle

\begin{abstract}
Federated learning is an approach to train machine learning models on the edge
  of the networks, as close as possible where the data is produced, motivated by
  the emerging problem of the inability to stream and centrally store the large
  amount of data produced by edge devices as well as by data privacy concerns.
  This learning paradigm is in need of robust algorithms to device
  heterogeneity and data heterogeneity. This paper proposes \texttt{ModFL} as a
  federated learning framework that splits the models into a
  \textit{configuration module} and an \textit{operation module} enabling
  federated learning of the individual modules. This modular approach makes it
  possible to extract knowlege from a group of heterogeneous devices as well
  as from non-IID data produced from its users. This approach can be viewed as
  an extension of the federated learning with personalisation layers
  \texttt{FedPer} framework that addresses data heterogeneity. We
  show that \texttt{ModFL} outperforms \texttt{FedPer} for non-IID data
  partitions of CIFAR-10 and STL-10 using CNNs. Our results on time-series data
  with HAPT, RWHAR, and WISDM datasets using RNNs remain inconclusive, we argue
  that the chosen datasets do not highlight the advantages of \texttt{ModFL},
  but in the worst case scenario it performs as well as \texttt{FedPer}.
\end{abstract}


\section{Introduction}
The collective amount of data created in modern life mobile devices, IoT home
devices, wearables and other edge devices is practically immeasurable. Often such large
amount of data makes it unfeasible to stream, collect and store the data for
further processing. On the other hand, traditional machine learning (ML)
approaches need to centrally collect the data for model training.  Additionally,
privacy concerns about data collection and government regulations, such as the
European GDPR, have set focus on federated learning~\cite{McMahan2016} as an
alternative way of training ML models that cope with such restrictions. This
approach trains the ML models close to the network edges and, therefore, close to
the data sources. Only the locally trained models are centrally collected and
aggregated to a global model. The \textit{De Facto} standard of federated
learning \texttt{FedAvg}~\cite{McMahan2016} averages the local weights from the
devices in the federation to a global model. This empirical approach has shown
to be stable even for non-convex optimisation problems and is used as
\textit{the benchmark} to compare to for newly developed federated learning
protocols.

Even though increasing research attention is shifting towards this field, it is
as yet at an early stage and core challenges are still to be solved.
The authors of \cite{Li2020} divide them into four categories; expensive communication, device
heterogeneity, data heterogeneity and privacy concerns. Our focus lies in
device and data heterogeneity.  Device heterogeneity comes from the fact
that different edge devices have different hardware and thus a variability in
CPU, memory, sensors or network connectivity, leading to a variability of input
features, of local training times or communication dropouts. One factor that often is not mentioned as part
of this variability is device generation or device model, \textit{i.e.}, edge
devices that perform the same task or measure the same observable by collecting
different types of inputs. For example, early versions of human activity
recording gadgets might only use the accelerometer for classifying the
activities, while later versions might have added a gyroscope sensor to improve
accuracy. This additional sensor is a challenge to traditional federated
learning approaches, as it needs to train from scratch a new model architecture.
In the present work we will focus on the generation difference type of device heterogeneity,
although this approach can be also used to ease the problem of different training
times due to computational power. Data heterogeneity arises as different users (edge devices) in a
federation, do in general generate data in a non-independent and non-identically
distributed (non-IID) way.  In the case of human activity classification,
different persons do spend different amount of time walking, running, jogging,
jumping, lying or sitting.  Some of them might never run or jump, while others
might have a regular training schedule. 

Recent work addressing the data heterogeneity has proposed
federated learning with personalisation layers
(\texttt{FedPer})~\cite{Arivazhagan2019}. In this approach, the authors propose
to divide the neural networks (NN) into so called base layers and
personalisation layers. This approach combats effectively the ill-effects of
data heterogeneity. From~\texttt{FedPer} we borrow
the notion of splitting the NN in two modules; \textit{a configuration module}
and an \textit{operation module}. We assume a scenario where different models of
edge devices have a different configuration of inputs (it can be number and/or
different kind of sensors) to perform the same task (measure the same
observable). At the same time, the edge devices are owned by different clients
that can be categorised by how they operate the edge devices.  Our main
assumptions are that the configuration modules of models trained on such devices
can be trained in a federated learning way between the devices of the same kind,
whereas the operation modules can be trained between the same group of operational 
clients. We call this concept modular federated learning \texttt{ModFL}. 

In this work we show that the \texttt{ModFL} framework using convolutional
neural networks (CNN) for image classification have a higher accuracy than the
\texttt{FedPer} approach. We show how the proposed \texttt{ModFL} protocol can
benefit of federation among like-minded clients to improve accuracy even for
non-IID data. Moreover, \texttt{ModFL} handels device generation heterogeneity
by grouping federation groups of same generation devices. For recurrent neural
networks (RNN) and time-series data, the results are not conclusive. We argue
that more challenging time-series data sets have the potential to show similar
accuracy boosts as for the CNN and image classification. 

The paper is structured as following: the next section summarises related work
to \texttt{ModFL}, following by necessary definitions and  describes
the~\texttt{ModFL} protocol proposed in this work, then we describe the
experiments and results that demonstrate the effectiveness of this protocol.
Finally, we give final conclusions and further work in the last section. 

\section{Related Work}
\label{sec:related_work}
Modularity in NN was investigated by~\cite{Watanabe2018} where the authors
showed that layered NN can be decomposed in small sets of independent neural
networks. They detected units of similar connection patterns. Recent work
from~\cite{Filan2021} shows that multi-layer perceptrons (MLP) neural networks
are significantly more modular than random networks with the same distribution
of weights when trained with weight pruning.  They show that this is not a
feature of the datasets it was trained on, but a general property of MLP. The current
work is motivated by these results; one question that proceeded the
development of \texttt{ModFL} framework is how to use this modularity property
to train individual sub-modules or sets of independent NN in a federated
learning fashion. 

The \textit{De Facto} standard of federated learning is federated averaging
(\texttt{FedAvg}) first introduced in the seminal work by \cite{McMahan2016}.
Here each device optimises the local version of the global model by stochastic
gradient descent with the same learning rate and number of local epochs. The
resulting local model updates are centrally averaged to a global model. It is
important to level the hyperparameters of \texttt{FedAvg} to allow as many local
epochs as possible to reduce communication cost, on the other hand these local
epochs cannot be too long as the local models can reach local optima deviating
too much from the global optimum, making convergence slower or even making the
method diverge. A proposed framework to reduce communication and
increase the local epochs while ensuring convergence is
\texttt{FedProx}~\cite{Li2018}, here a proximal term is added to the cost
function to be optimised during the learning period. This term helps the local
updates stay closer to the global model enabling larger local epochs and thus
minimising communication costs. Furthermore, a variable number of local
iterations across different edge devices is allowed, making it possible to
adjust this number to the device hardware capabilities.  

To address data heterogeneity in a federated learning setup,
\textit{i.e.}, differences in the data sources, the \texttt{FedPer} framework
proposed to split the neural networks into \textit{base layers} and
\textit{personalisation layers}~\cite{Arivazhagan2019}. The latter ones even
after local training remain in the edge devices and are not shared, whereas the
former ones will follow collective training frameworks such as the
\texttt{FedAvg}. In this way each device will have a unique model composed of
the \textit{global} base layers and the \textit{local} personalisation layers.
The assumption here is that the personalisation layers learn the specific
features of the data coming from the personal users habits, in contrast the base
layers learn more general features that are common to all users. By splitting
and training the network in such a way, the trained models perform better in
non-IID data compared to \texttt{FedAvg}. In this work, the proposed framework of
\texttt{ModFL} extends the \texttt{FedPer} to also deal with device heterogeneity
of edge devices. 

Further approaches to personalised federated learning that tackle the
heterogenous data problem are \texttt{Per-FedAvg} that uses meta-learning to
train a global model that has ``good'' initial values such that users can
subsecuently adapt the model, within limited computational expences, to better
generalise to the user's data~\cite{Fallah2020}; adaptive personalised federated
learning \texttt{APFL} proposes a mixture of the local and the global model,
where each client trains its local model while contributing to the global model.
The personalised model that adapts better to the user's data emerges from the
optimization of a convex combination of the local and global
model~\cite{Deng2020}. These approaches tackle the data heterogeneity problem by
making use of a global model to later derive a personalised model, indirectly
assuming a common model architecture between all users, thus not taking into
account device heterogeneity. This lies in contrast to the proposed approach
here, where the model itself is split into ``modules'' that are federated
within their common classes, allowing to have different model architecture and
size per client by combining different ``modules''. A recent approach that uses
clustering between similar users is \texttt{PerFed-CKT}~\cite{Cho2021}. Here the
authors propose clustered co-distillation where clients transfer their knowlege
to other clients that have similar data-distribution. Because in this setup
logits are shared instead of model parameters clients with different model
architectures and sizes can federate between each other, thus tackling the device
and data heterogeneity problem. This approach uses a common dataset that is
shared between all clients, the clustering is based on the similarity of the
models output of the common dataset. Similar to the present work,
\texttt{PerFed-CKT} tackles the device and data heterogeneity problem, but differs
in that \texttt{PerFed-CKT} does need a common (public) data set shared between
all clients, whereas \texttt{ModFL} does not need such a public dataset. Another
difference is the global model aggregation, \texttt{ModFL} averages over the
model parameters, wheras \texttt{PerFed-CKT} does averages over the logits of
the clients models output based on the public dataset.

%
\begin{figure*}[t]
  \centering
  \includegraphics[width=\textwidth]{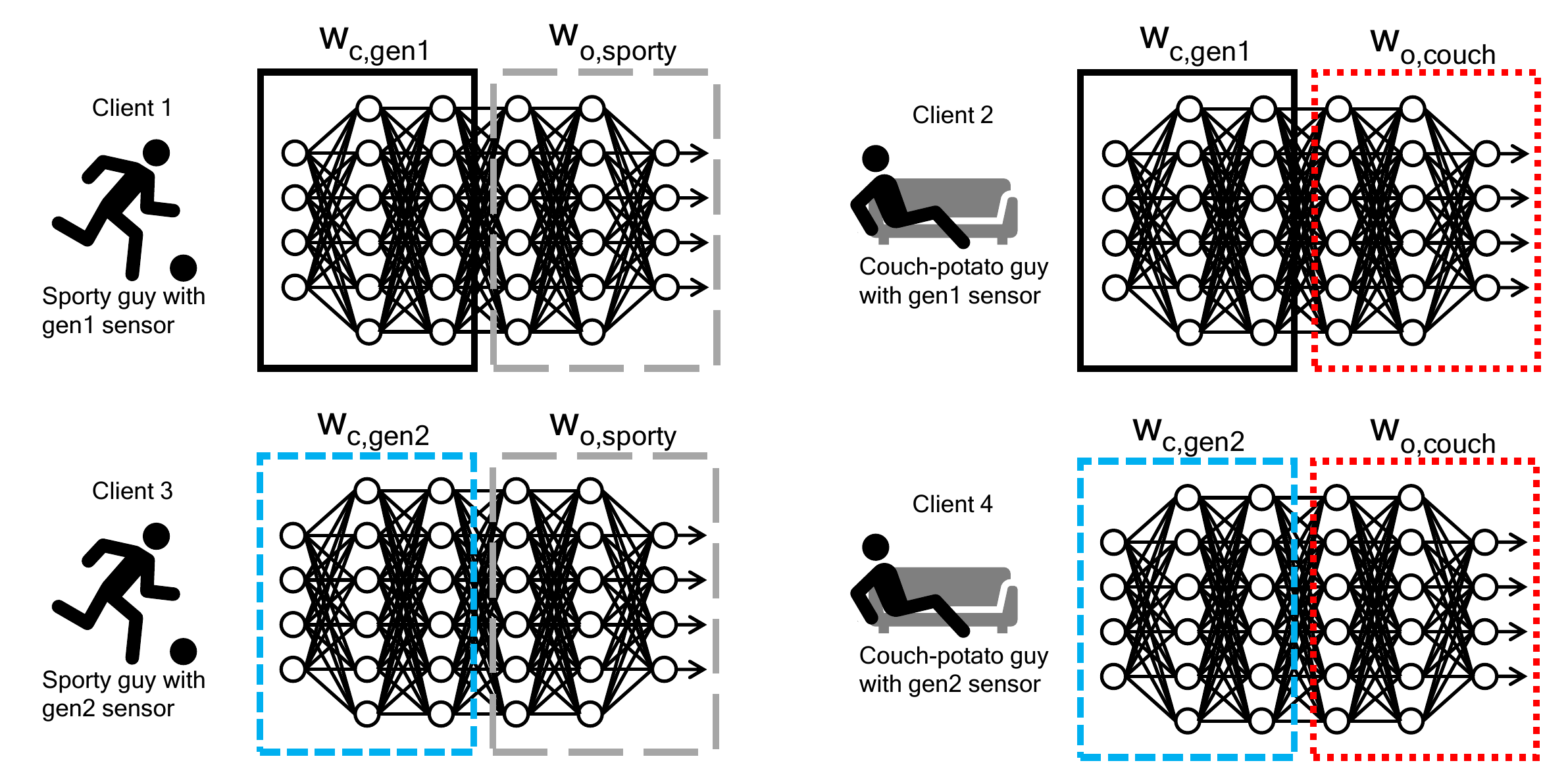}
  \caption{Main idea of \texttt{ModFL}, where the neural network models are
  split into a configuration module and an operation module and federated with
  similar cohorts denoted by the weight notations and boxed with various
  line styles and colors.
  \label{fig:modfl}
    }
\end{figure*}

\section{ModFL Framework}
\label{sec:modfl}

Modular federated learning (\texttt{ModFL}) is inspired 
by Scania's modular system~\cite{Skold2017} and 
by modularity in neural
networks~\cite{Watanabe2018,Filan2021} combined with federated learning with
personalisation layers \texttt{FedPer}~\cite{Arivazhagan2019}. For this \texttt{ModFL}
concept, consider first a deep neural network model $M_1$ with many
layers where we divide the model into two modules; a configuration module
$W_{c,1}$ consisting of the input layer with some following hidden
layers and an operation module $W_{o,1}$ consisting of the rest of the
hidden layers and the output layer. Similar to \texttt{FedPer}, the configuration module 
$W_{c,1}$ will be federated among all clients with model $M_1$.  Now consider another model $M_2$
that performs the same task as $M_1$ but for a different generation of
edge device, and similarly divide the model into a configuration module
$W_{c,2}$ and a operation module $W_{o,2}$. We have the
assumption that different generation edge devices, even though they perform the
same task, \textit{i.e.}, they measure the same observable, generate different
type of data. Therefore, $W_{c,1}$ and $W_{c,2}$, are in
general different in terms of neurons and layers to accommodate the previously
discussed differences.  However, the operation modules, $W_{o,1}$ and
$W_{o,2}$, have to have the same architecture so that they are able to do
federated learning between each other, \textit{i.e.} $W_{o,1} =
W_{o,2} = W_o$. If all clients would operate in similar
ways, \textit{i.e.}, the data is IID in all clients, then all
the operation modules $W_o$ from the different clients could federate
between each other. However, in general in a federated learning setup, the clients
generate non-IID data. To deal with this, the authors of \cite{Arivazhagan2019} introduced the
personalisation layers, which would be local to every client and would not
federate. In \texttt{ModFL}, we assume that client operations
can be categorised, whether from a priori knowledge of the edge device usage
(operation) or by a similarity measure such as the one proposed by
\cite{Sattler2019}. To simplify the setup, we assume that a priori
knowledge of the clients is available. \texttt{ModFL} then proposes to federate the operation modules
between clients belonging to the same operation category cohort. This way the configuration modules of 
\texttt{ModFL} deal with device heterogeneity while the operation modules deal with data heterogeneity. 

The main idea of \texttt{ModFL} is to federate the different configuration and
operation modules of the deep neural network with its corresponding peers. This
means that the configuration module will not necessary federate with the same
peers as the operation module. To illustrate it with an example, we
assume that we have four clients, each of them will train a local model
$M^{(n)}=W_c^{(n)}+W_o^{(n)}$ where client $n \in \left\{1,2,3,4\right\}$ and 
the addition sign $(+)$ represents concatenation. 
Two types of edge device generations for human activity categorisation are 
used by the clients; gen1 has only one sensor for categorising the activities 
and gen2 has two sensors. Therefore, we have two different model architectures
$\left\{M_{gen1},M_{gen2}\right\}$. Clients $C_{gen1}=\left\{1,2\right\}$ use gen1 devices and 
clients $C_{gen2}=\left\{3,4\right\}$ gen2. Then $M^{(n)}=W^{(n)}_{c,gen1}+W_o^{(n)}\vert_{n=1,2}$ and
$M^{(n)}=W^{(n)}_{c,gen2}+W_o^{(n)}\vert_{n=3,4}$. Additionally,
the clients $O_{couch-potato}=\left\{2,4\right\}$ are known to belong to the "couch-potato" cohort and the clients
$O_{sporty}=\left\{1,3\right\}$ to the "sporty" cohort. Therefore, the local models are composed of
different configuration modules and operation modules:
\begin{itemize}
	\item $M^{(1)}=W^{(1)}_{c,gen1}+W^{(1)}_{o,sporty}$,
	\item $M^{(2)}=W^{(2)}_{c,gen1}+W^{(2)}_{o,couch-potato}$,
	\item $M^{(3)}=W^{(3)}_{c,gen2}+W^{(3)}_{o,sporty}$,
	\item $M^{(4)}=W^{(4)}_{c,gen2}+W^{(4)}_{o,couch-potato}$,
\end{itemize}
see Fig.~\ref{fig:modfl} for clarity of this example. In this simple scenario the model $M^{(1)}$
will benefit from learning the features of the gen1 edge device from the
federation with user $n=2$, simultaneously model $M^{(1)}$ will benefit to improve
the classification performance of the human activities relevant for a sporty
user by federating with user $n=3$. Note that the \texttt{ModFL} concept is not limited to only two 
configuration groups and operation groups but can be extended to multiple groups. 

\begin{algorithm}[t]
	\caption{\texttt{ModFL} algorithm}\label{alg:modfl}
	\begin{center} \textbf{\texttt{The n-th client}} \end{center}
	\begin{algorithmic}[1]
		\REQUIRE data $D^{(n)}$
		\STATE Recieve $(W_c, W_o)$ from server
		\STATE $(W_c, W_o) \gets \textrm{gradient descent} (W_c, W_o, D^{(n)})$ 
		\STATE Send updated $(W_c, W_o)$ to server
	\end{algorithmic}
	
	\begin{center} \textbf{\texttt{The server}} \end{center}
	\begin{algorithmic}[1]
		\REQUIRE Set of models $\left\{M_1,...,M_m\right\}$ corresponding to the different client configurations, models split point $s$.
		\FORALL {$M_i$}
			\STATE $(W_{c,i}, W_{o}) \gets$ split($M_i$) based on $s$
		\ENDFOR
		\STATE Initialise all $W_{c,i}$ and $W_{o}$ at random
		\STATE Cluster clients into corresponding configuration groups $\left\{C_1,...,C_m\right\}$ such that there is a one-to-one mapping $\left\{C_i\right\}\rightarrow\left\{W_{c,i}\right\}$ for $i\in\left\{1,...,m\right\}$
		\STATE Send $(W_{c,i}, W_{o})$ to the corresponding clients according to the mapping
		\FOR {$k=1,2,...$}
			\STATE Receive $W^{(n)}_{c,i}$ and $W^{(n)}_{o}$ from all clients
			\FOR {$i \in \left\{1,...,m\right\}$}
				\STATE $W_{c,i} \gets \overline{W^{(n)}_{c,i}}\;\; \forall n \in C_i $
			\ENDFOR
			\STATE Cluster clients into operational clusters $\left\{O_1,...,O_l\right\}$, either by a-priori operational knowlege or using a clustering method on the operational modules $W^{(n)}_{o}$ and label them ${W^{(n)}_{o,j}}$ such that there is a one-to-one mapping $\left\{O_j\right\}\rightarrow\left\{W_{o,j}\right\}$ for $j\in\left\{1,...,l\right\}$
			\FOR {$j \in \left\{1,...,l\right\}$} 
				\STATE $W_{o,j} \gets \overline{W^{(n)}_{o,j}} \;\; \forall n \in O_j$
			\ENDFOR
			\STATE Send the global configuration modules $W_{c,i}$ to the corresponding clients in $C_i$
			\STATE Send the global operational modules $W_{o,j}$ to the corresponding clients in $O_j$
		\ENDFOR
	\end{algorithmic}
\end{algorithm}

The detailed and generalized algorithm of \texttt{ModFL} is shown in Algorithm~\ref{alg:modfl}.
Each client gets weights from the server, updates the weights based on their own
data $D^{(n)}$  with a gradient descent algorithm and sends back the updated
weights to the server. This continues in many communication rounds until
convergence or requirements are met. The server initialises the models $M_i$
according to the number of different configuration or generations and splits the
models to configuration and operation modules $(W_{c,i}, W_{o})$. The
corresponding weights are sent to the clients with the corresponding
configuration group $C_i$ for training. For each step, the server
receives updated weights $(W^{(n)}_{c,i}, W^{(n)}_{o})$ from all clients. For all the
different configuration groups $\left\{C_i\right\}_{i=1,...,m}$, average the weights
$W^{(n)}_{c,i}\vert_{n\in C_i}$ from the clients belonging to that group. Similarly for operation
groups $\left\{O_j\right\}_{j=1,...,l}$ with weights $W^{(n)}_{o,j}\vert_{n\in O_j}$. Lastly, send all the
aggregated modules $\left\{W_{c,i}\right\}$ and $\left\{W_{o,j}\right\}$ to the corresponding 
clients in  $\left\{C_i\right\}$ and $\left\{O_j\right\}$, respectively.

\texttt{ModFL} is not limited to a single task problem with one and same 
operation module $W_o$. It can be extended to multitask learning problems 
\cite{Crawshaw2020}, where the operation module models can be different 
for different tasks, \textit{i.e.} each task can be represented with different 
neural network model for the operation module. For example using the same 
setting in the above example, one task could be to detect injury while sporting 
for the sporty cohort and another task could be to predict falls for the couch-potato
cohort. These two different tasks will in general be different models and will be
reflected in the operation module. As a reminder, configuration modules for gen1 
and gen2 are in general different models to accommodate the differences of the 
component generations. The computational complexity for \texttt{ModFL} does not 
change on the client side compared to \texttt{FedAvg} as it does the same local training. 
However, on the server side, the computational complexity remains if a-priori operational
knowledge is known (row 12 in Algorithm~\ref{alg:modfl}), otherwise it may increase 
depending on the clustering method. 

\section{Experimental Setup}
\label{sec:experiment}
In order to show the concept of \texttt{ModFL}, at least two different datasets
with similar tasks are required. We evaluate the performance of \texttt{ModFL}
with two different classification tasks, namely image and human activity
classification, using CNN and RNN models, respectively. Additionally, the
data is non-identical partitioned on the labels among operation groups, thus
tuning the degree of data heterogeneity. We will benchmark the results
with federated learning with personalisation layer \texttt{FedPer} and the 
vanilla federated learning \texttt{FedAvg}. 

\subsection{Datasets}
\paragraph{Image classification}
CIFAR-10\footnote{https://www.cs.toronto.edu/~kriz/cifar.html}~\cite{Krizhevsky2009}
and STL-10\footnote{https://cs.stanford.edu/~acoates/stl10/}~\cite{Coates2011}
are both image classification datasets with 10 class labels with different image
resolutions, $32 \times 32$ and $96 \times 96$\,pixels, respectively and each image has a unique
label. Both datasets have 9 labels in common, which makes this suitable to test
\texttt{ModFL}. Due to the difference in the number of data samples, we evened the amount of 
data to 11700 data samples per dataset with a 75/25 train and test ratio. 

\begin{figure}[t]
     \centering
     \includegraphics[width=\columnwidth]{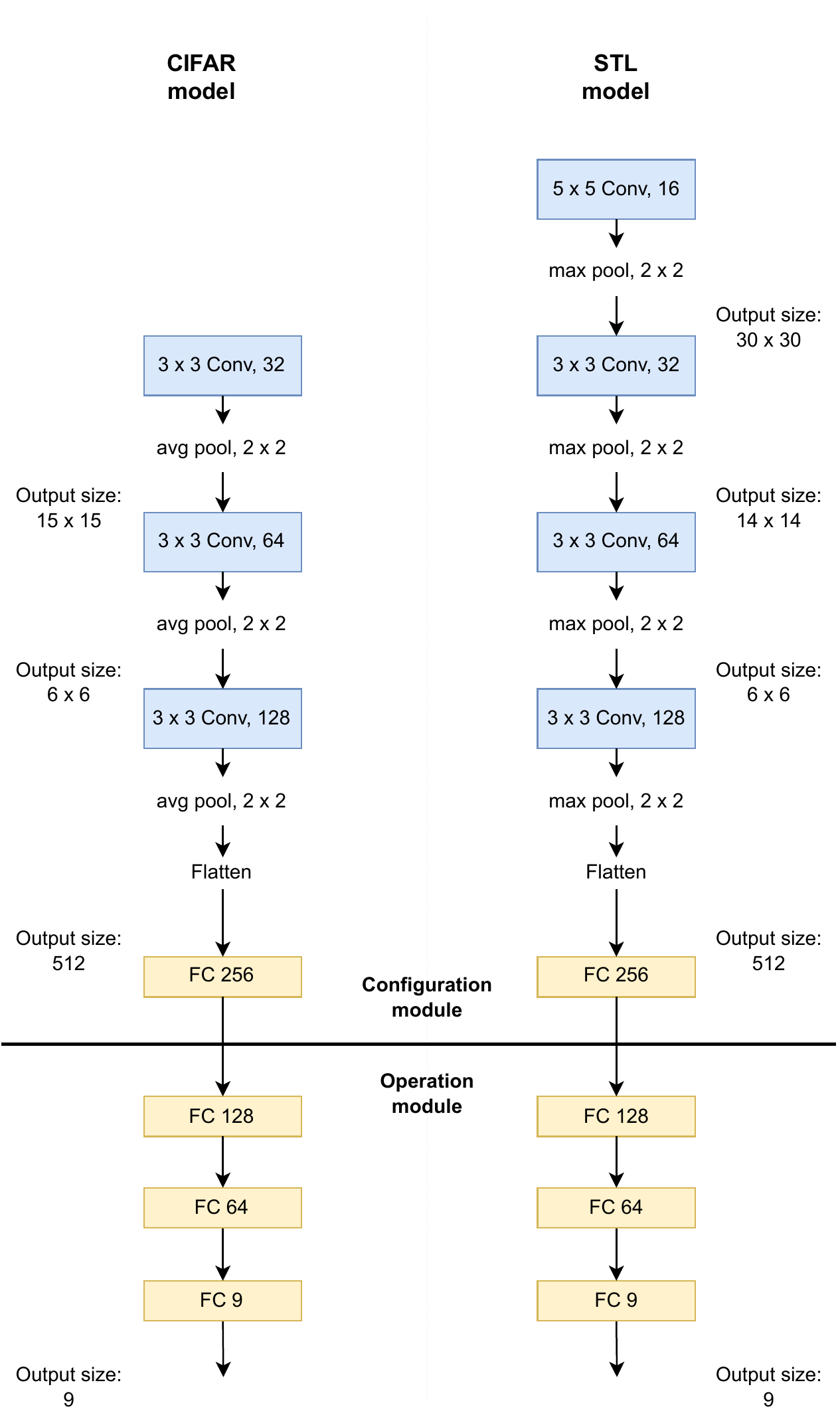}
     \caption{Image classification models. }
     \label{fig:img_models}
\end{figure}

\paragraph{Human activity classification}
HAPT\footnote{http://archive.ics.uci.edu/ml/datasets/Smartphone-Based+\\Recognition+of+Human+Activities+and+Postural+Transitions}~\cite{Reyes-Ortiz2016},
WISDM\footnote{https://www.cis.fordham.edu/wisdm/dataset.php}~\cite{Kwapisz2011},
and RealWorld
HAR\footnote{https://sensor.informatik.uni-mannheim.de/\#dataset\_realworld}~\cite{Sztyler2016}
(RWHAR) are three different human activity recognition datasets, in which they
have 5 labels in common. Each data sample has a unique label. WISDM collects
accelerometer time-series data from smartphones with a frequency of 20\,Hz,
while HAPT and RWHAR collect accelerometer and gyroscope time-series data with
a 50\,Hz frequency. Although RWHAR has sensors in different locations (head, chest, arm, waist and more), only the one from the waist was
used as that is similar to WISDM. Due to the difference in the number of data
samples for each label and dataset, we evened the amount of data to 1250 data samples
per dataset with a 80/20 train and test ratio.

\subsection{Model Architectures and Hyperparameters}
We designed our own models to evaluate \texttt{ModFL}. Recall that the configuration 
modules can be different to handle different datasets while the operation modules 
have to be the same for the same task, in this case for classification, across the models. 
Our aim is not to get the highest test accuracy with these proposed models, but rather
show that the \texttt{ModFL} concept can improve it while addressing device and data 
heterogeneities. 

\paragraph{Image classification} For the image classification, we use CNN layers
followed by fully connected layers. Since CIFAR-10 and STL-10 have different
resolution images, we designed the configuration module layers differently while
the operation module layers remain the same. For the CIFAR model, we designed
the configuration module with 32--64--128 units of 2D convolutional layers with
average pooling in between followed by a 256 units fully connected layer, in
total 4 layers. The configuration module for STL model is designed with
16--32--64--128 units of 2D convolutional layers with max pooling in between
followed by a 256 units fully connected layer, in total 5 layers. The
operation module for both models are 128--64--9 units of fully connected layers.
See Fig.~\ref{fig:img_models} for an illustration of the models.
All layers used ReLU (rectified linear unit) as activation function except the
last fully connected layer which used softmax. 

\begin{figure}[t]
     \centering
     \includegraphics[width=\columnwidth]{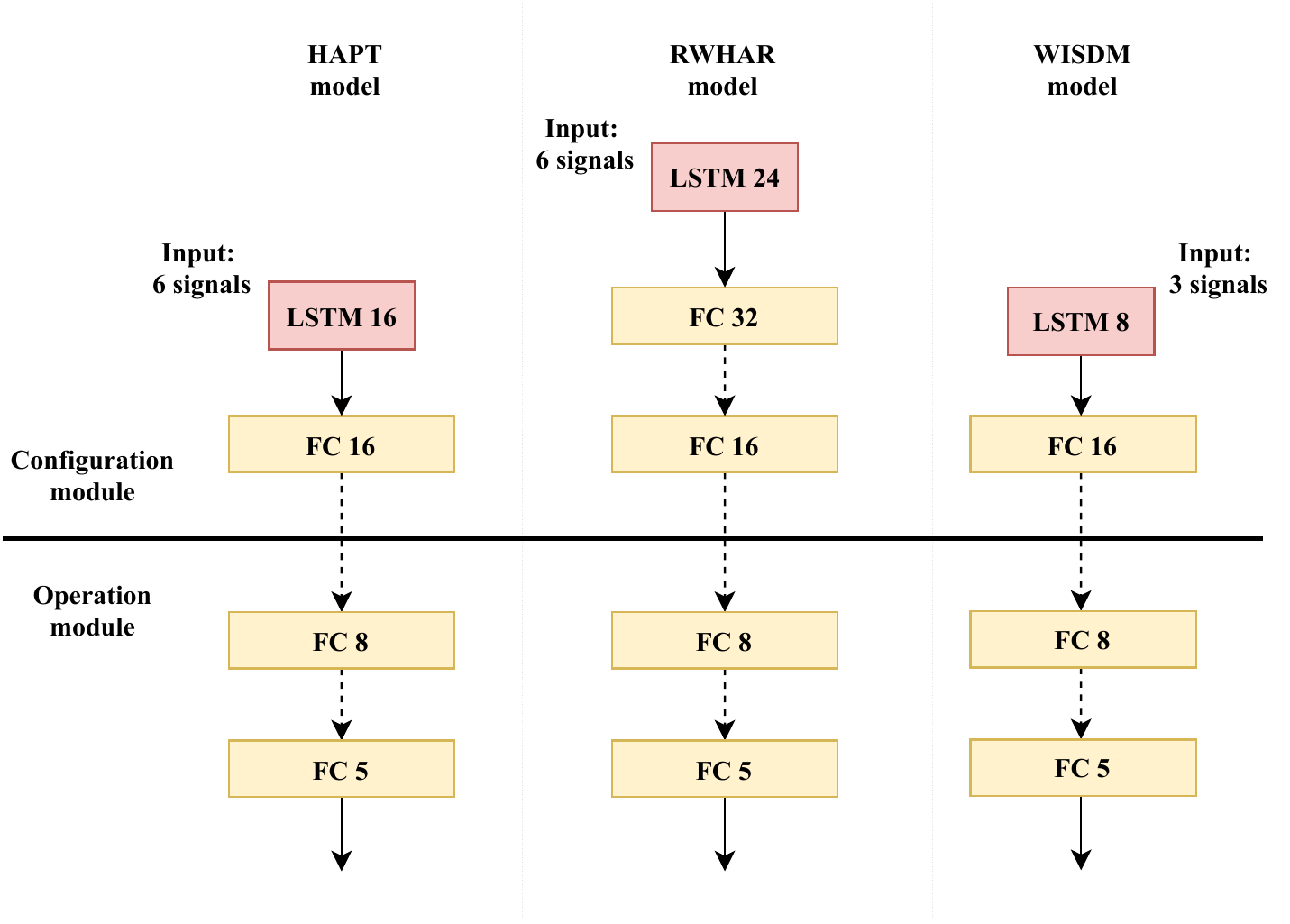}
     \caption{Human activity classification models. }
     \label{fig:har_models}
\end{figure}

\paragraph{Human activity classification} Since the human activity classification
datasets are time-series data, the natural way for classification is to use RNN,
for this case we used long-short term memory (LSTM), with fully connected layers. For the
HAPT model, we designed the configuration module with a 16 units LSTM layer
followed by a 16 units fully connected layer. For the RWHAR model, we used a 
24 units LSTM layer with 32--16 units of fully connected layers. Lastly, for the 
WISDM model we had an 8 units LSTM layer with a 16 units fully connected layer. 
The operation module consisted of 16--5 units of
fully connected layers. See Fig.~\ref{fig:har_models} for an illustration of
the models. All layers used tanh (hyperbolic tangent) as activation function except the
last fully connected layer which used softmax. 

For the federated learning training to converge easier, the weight
initialisation should be the same for all clients~\cite{McMahan2016}. In our
case, the configuration module layers from the same model have to be initiated
with the same weights and the operation module layers for all clients have to be
initiated with the same weights. Similarly to the conclusions
from~\cite{Arivazhagan2019} where different models have slightly better test
accuracy with different amount of personalised layers, we conclude that the cut
between the configuration module and the operation module of a model is an
additional hyperparameter to be tuned for each specific case. 

We use Adam optimiser with a learning rate of 0.001. For training the image
classification, we set local epochs to 1, the global communication rounds to
200, and a batch size of 16. For training the human activity classification, we
set local epochs to 1, the global communication rounds to 2000, and a batch size
of 8. The server receives all the incoming weights, divides them into configuration
and operation modules, and computes federated averaging
\texttt{FedAvg}~\cite{McMahan2016} accordingly to the groups. 

\begin{figure}[t]
	\centering
	\includegraphics[width=\columnwidth]{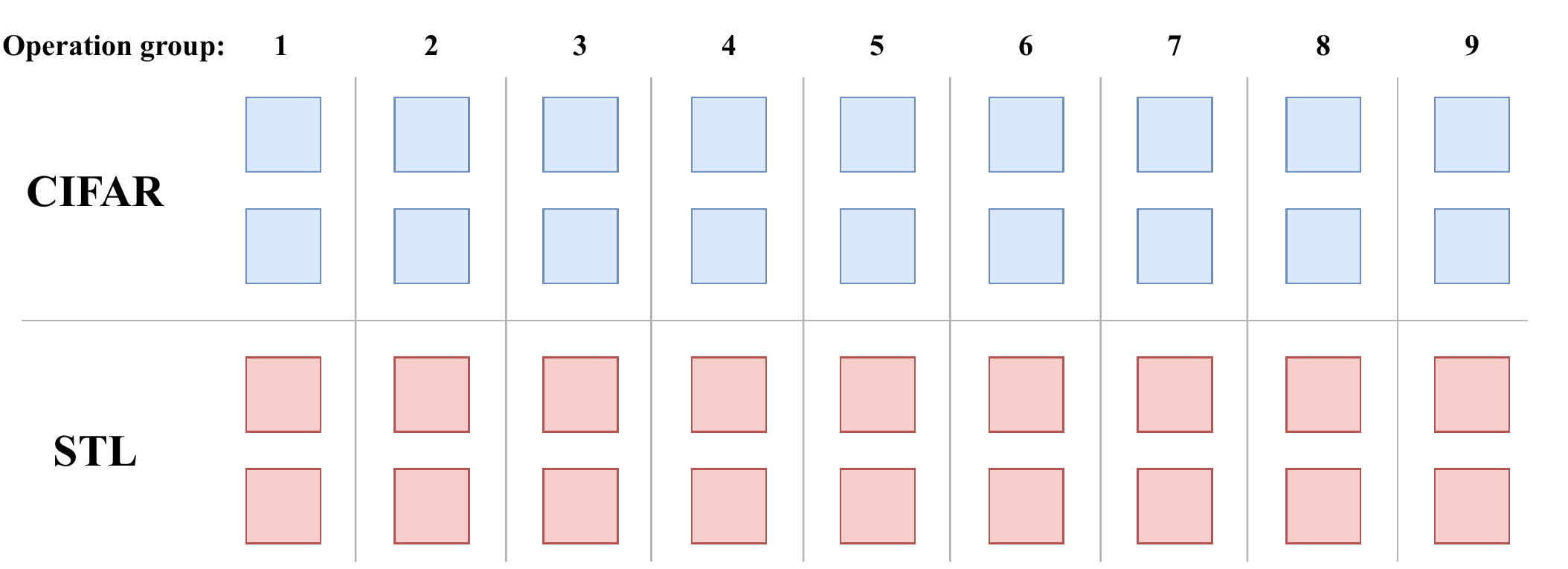}
  	\caption{Example of client setup with ${N}=36$ clients, where 18 clients use CIFAR model and the other 18 clients the STL model (\textit{i.e.} $|\{C_i\}|=2$) and $|\{O_j\}|=9$ different operation groups with 4 clients in each group.}
	\label{fig:clientsetup}
\end{figure}

\subsection{Client Setup}
In our setup, every client has the same amount of data and the same amount of data per given label. For the image
classification task the number of clients is ${N} \in \{18, 36, 54, 72\}$, half of
them use the CIFAR-10 dataset and model and the other half use the STL-10
dataset and model, \textit{i.e.} $|\{C_i\}| = 2$. We divide the the total number of clients into $|\{O_j\}| = 9$
different operation groups, \textit{i.e.}, if ${N}=36$, then 18 clients use the
CIFAR model and the other 18 the STL model. Additionally, these 36 clients are
divided into 9 groups (4 clients in each group), which will represent different
operations, see Fig.~\ref{fig:clientsetup} for clarity. Note that since we have limited amount of data for both tasks, the
more clients we have for training, the less data each client has. 

For the human activity classification task, the number of clients is ${N} \in
\{15, 30, 45, 60\}$. Since there three different datasets are used, $|\{C_i\}|=3$,
a third use HAPT model, another third the RWHAR model and
last third the WISDM model. We divide the total number of clients into $|\{O_j\}|
= 5$ different operation groups. 

Furthermore, we create three different levels of data heterogeneities
${P}$. For the image classification task, we chose ${P}_{img} \in
\left\{3,6,9\right\}$ unique labels per operation group and for the human
activity classification task we chose ${P}_{har} \in \left\{2,4,5\right\}$,
\textit{e.g.}, ${P}_{img} = 3$ means that each operation group has a set of 3
labels of unique combination. However, note that some labels will overlap with other
operation groups. ${P}_{img} = 9$ corresponds to all labels, which is an IID case,
and this means that all clients belong to one and the same operation group
($|\{O_j\}|=1$), similarily when ${P}_{har}=5$. This reduces the problem to
\texttt{FedAvg} on the operation module for all the clients. This special case
can also be seen as kind of reversed \texttt{FedPer}, the configuration module is
personalised to the generation of devices and the operation module is federated among all the
clients. 

Note that when we compare \texttt{ModFL} to \texttt{FedPer} or \texttt{FedAvg}, we split the
\texttt{ModFL} results into the respective dataset models (\textit{e.g.} CIFAR and STL models). This way, 
we can compare the differences since \texttt{FedPer} and \texttt{FedAvg} needs to have the
experiments separated (\textit{e.g.} one run with CIFAR model and dataset and another run with
STL model and dataset). 

\begin{figure}[t]
	\centering
	\includegraphics[width=\columnwidth]{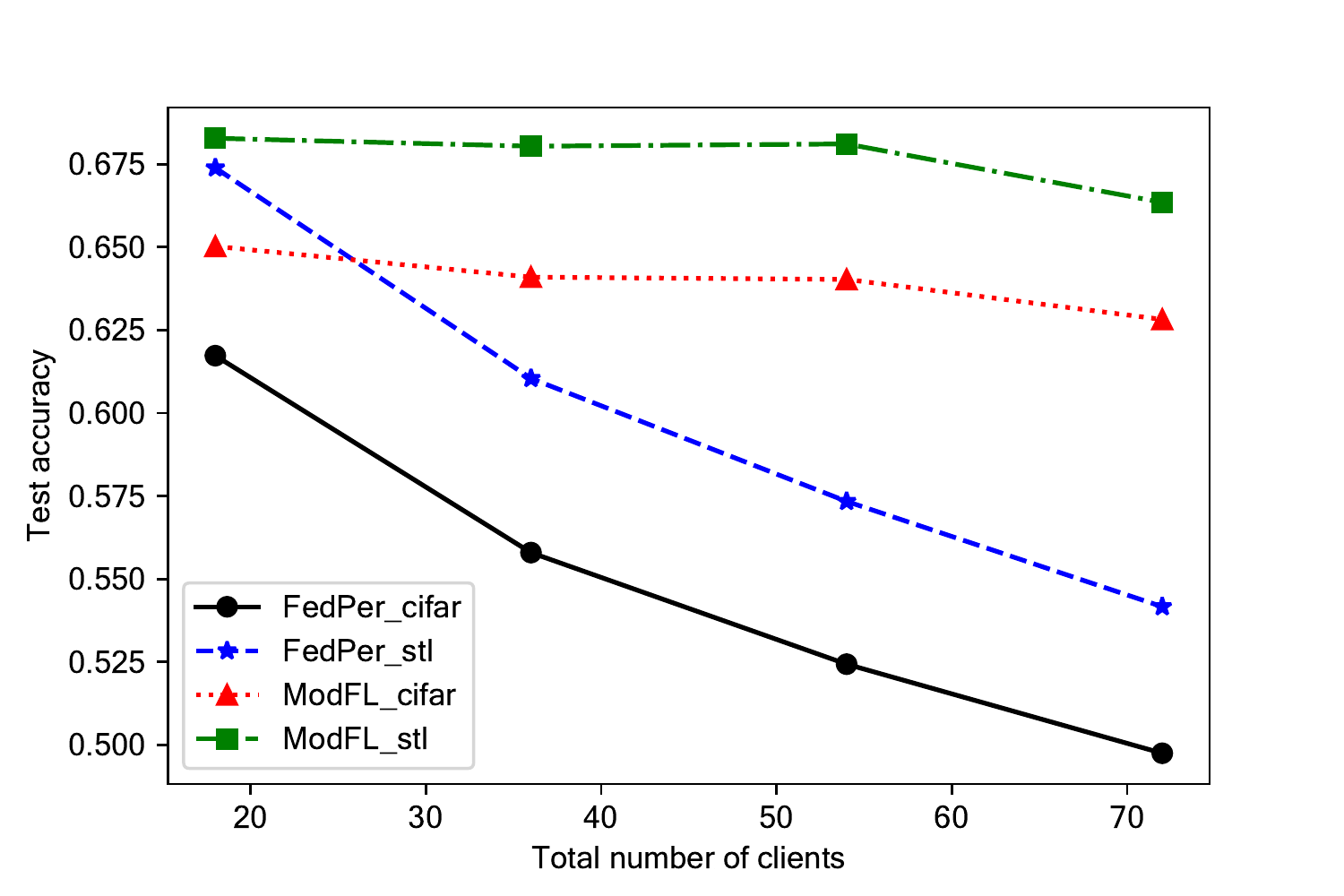}
  \caption{Test accuracy with respect to the number of clients on the image
  classification task with ${P}_{img}=6$. \texttt{ModFL} does not suffer as much
  compared to \texttt{FedPer} as the number of clients increases and the amount
  of data per client decreases.
  \label{fig:client_vs_acc} }
\end{figure}

\begin{figure*}[t]
  \begin{subfigure}[b]{0.32\textwidth}
	         \centering
        	 \includegraphics[width=\textwidth]{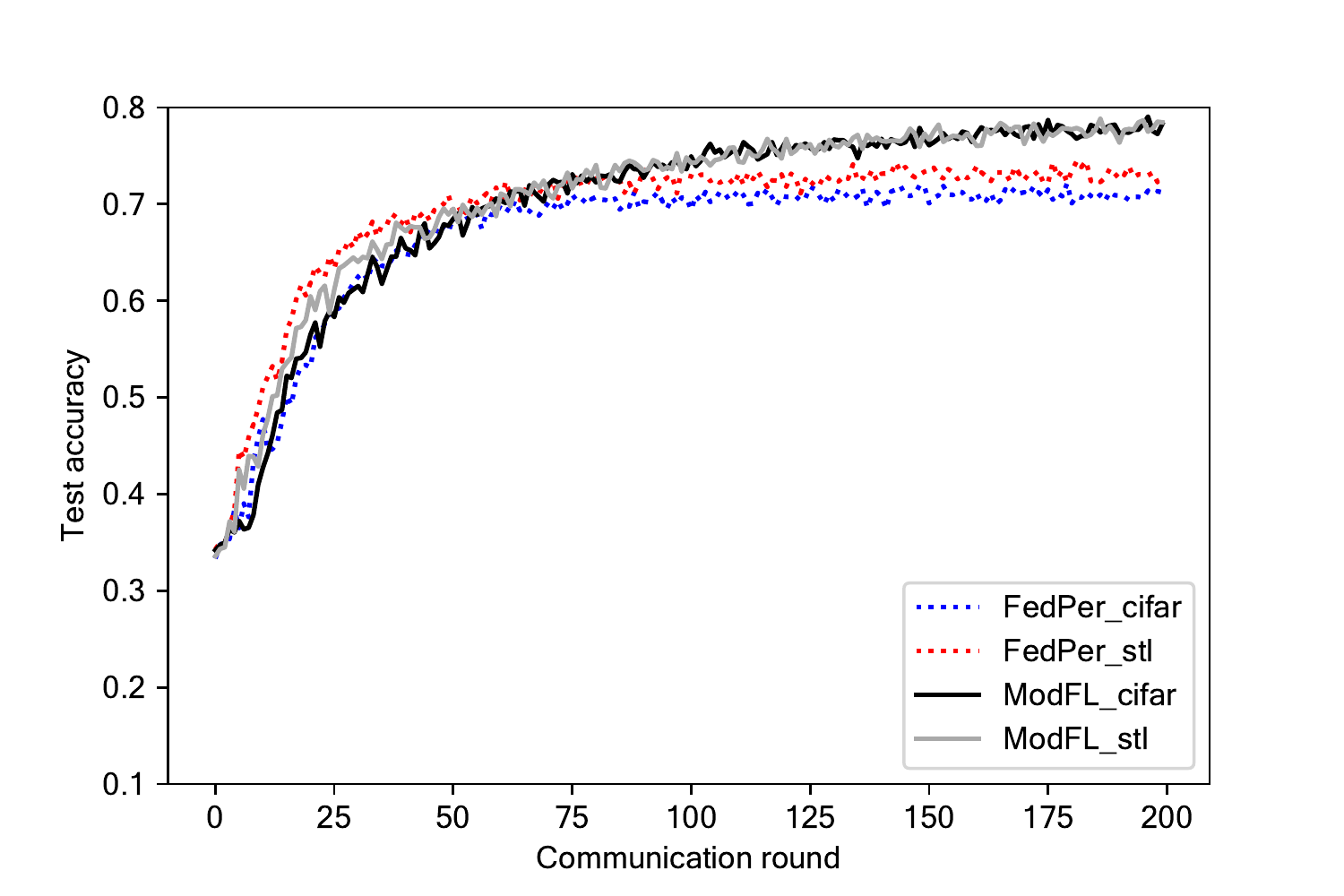}
		 \caption{}
         	 \label{fig:img_results1}
	 \end{subfigure}
	 \hfill
   \begin{subfigure}[b]{0.32\textwidth}
	         \centering
        	 \includegraphics[width=\textwidth]{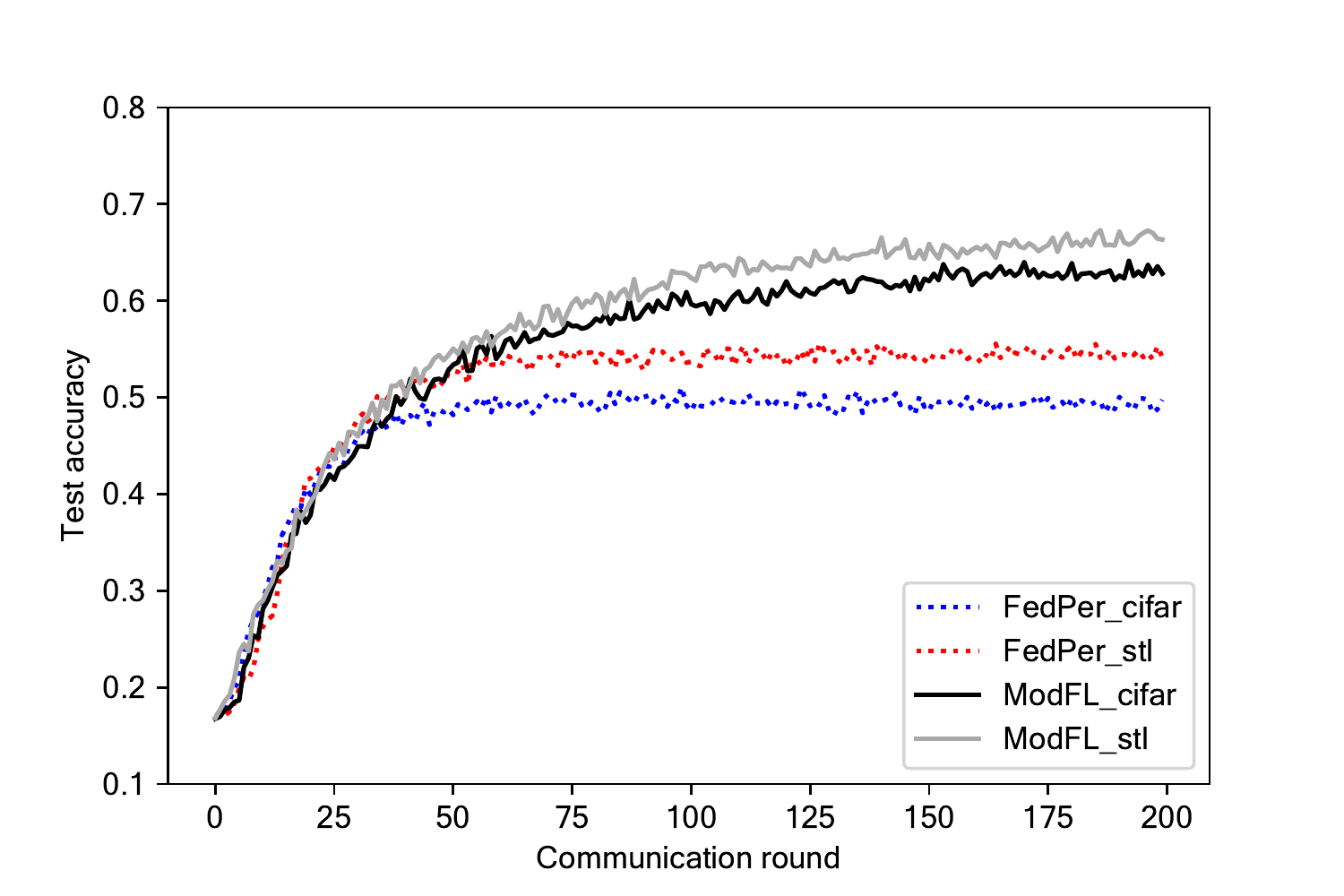}
		 \caption{}
         	 \label{fig:img_results2}
	 \end{subfigure}
	 \hfill
   \begin{subfigure}[b]{0.32\textwidth}
	         \centering
        	 \includegraphics[width=\textwidth]{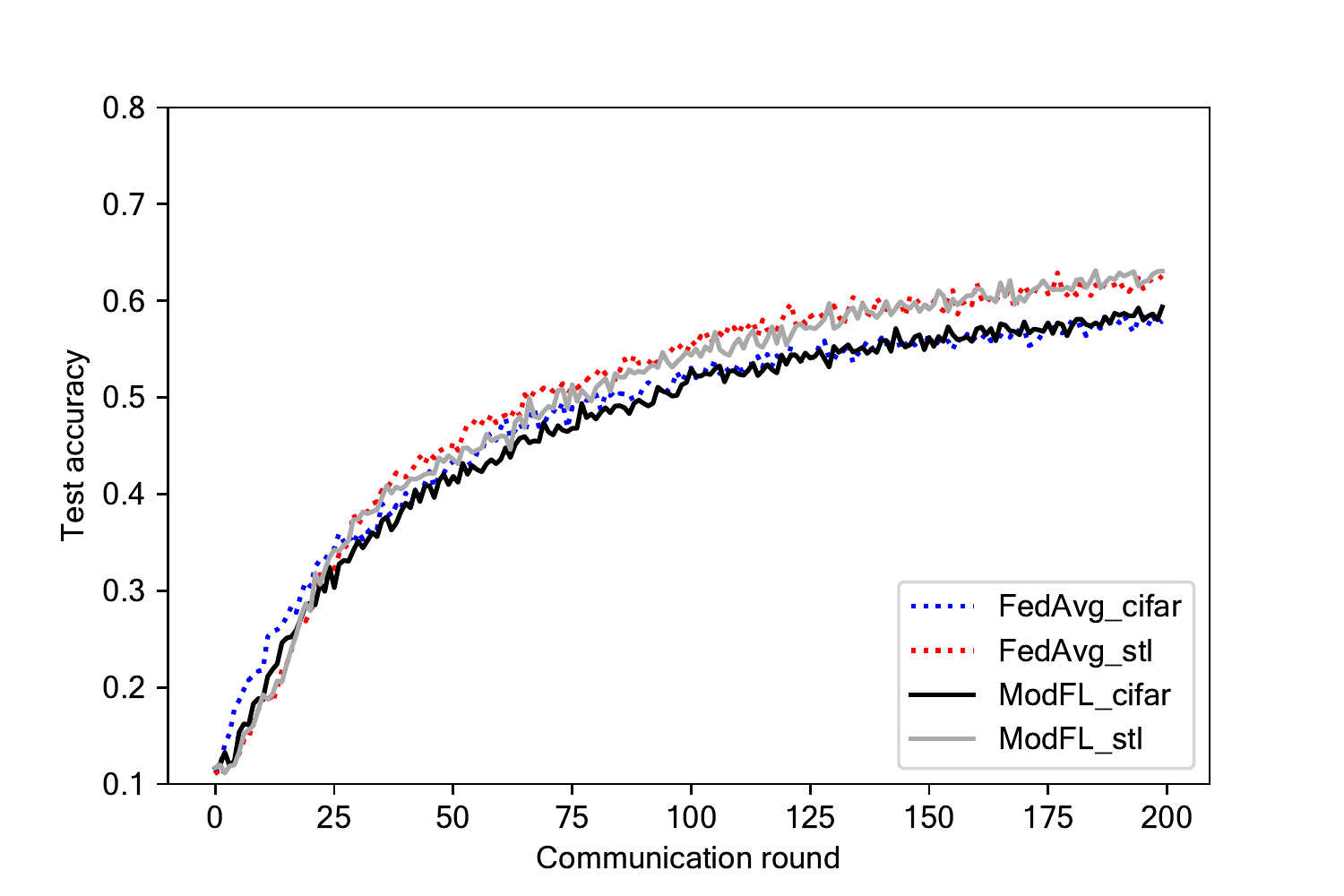}
		 \caption{}
         	 \label{fig:img_results3}
	 \end{subfigure}
     \caption{
       Performance of \texttt{ModFL} on image classification using CNN:
       (a) Non-IID partition with ${P}_{img}=3$ labels compared to \texttt{FedPer},
       (b) Non-IID partition with ${P}_{img}=6$ labels compared to \texttt{FedPer},
       (c) IID partition compared to \texttt{FedAvg}.
       \label{fig:img_results} }
\end{figure*}

\begin{figure*}[t]
	\begin{subfigure}{0.32\textwidth}
	         \centering
        	 \includegraphics[width=\textwidth]{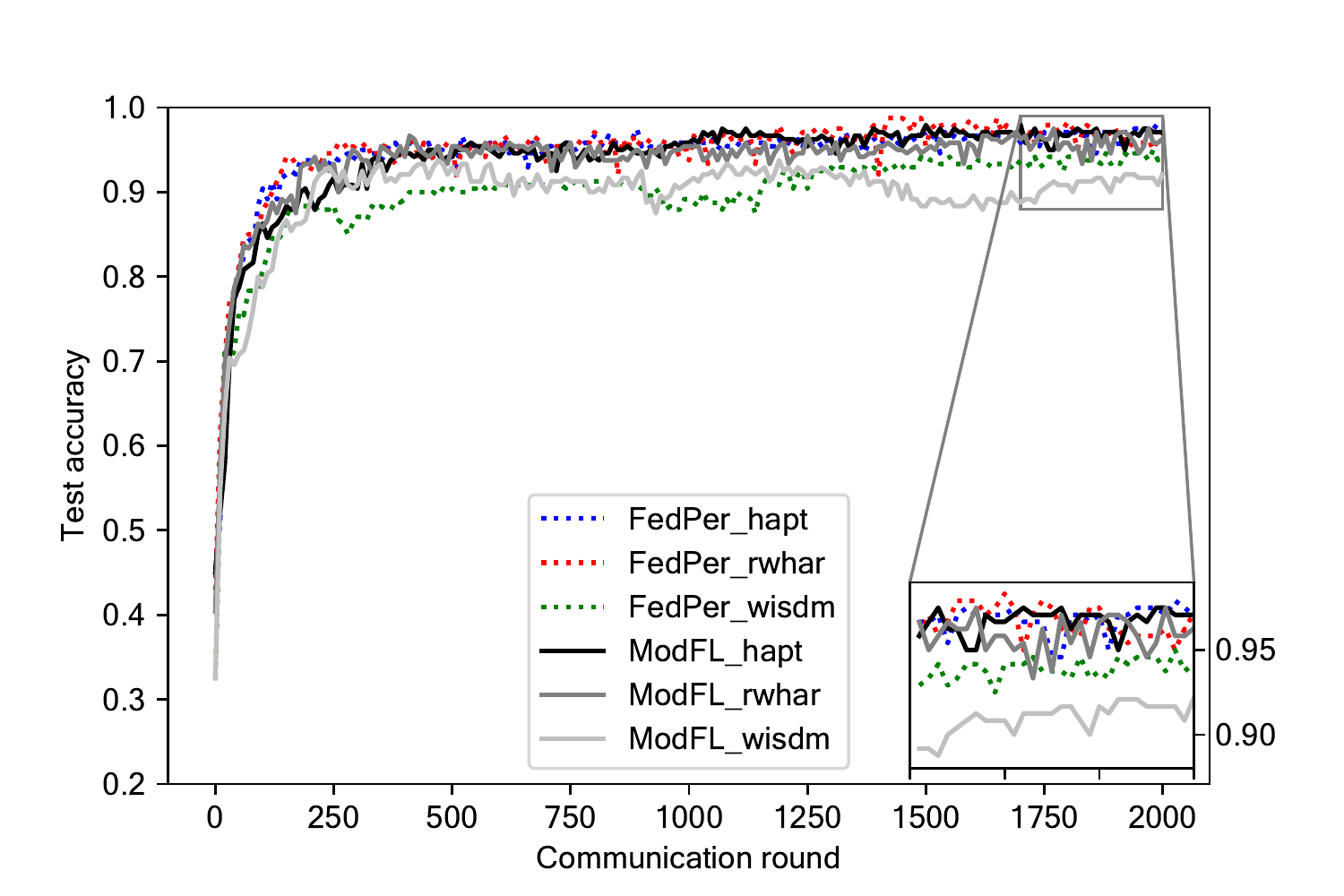}
		 \caption{}
         	 \label{fig:har_results1}
	 \end{subfigure}
	 \hfill
	 \begin{subfigure}{0.32\textwidth}
	         \centering
        	 \includegraphics[width=\textwidth]{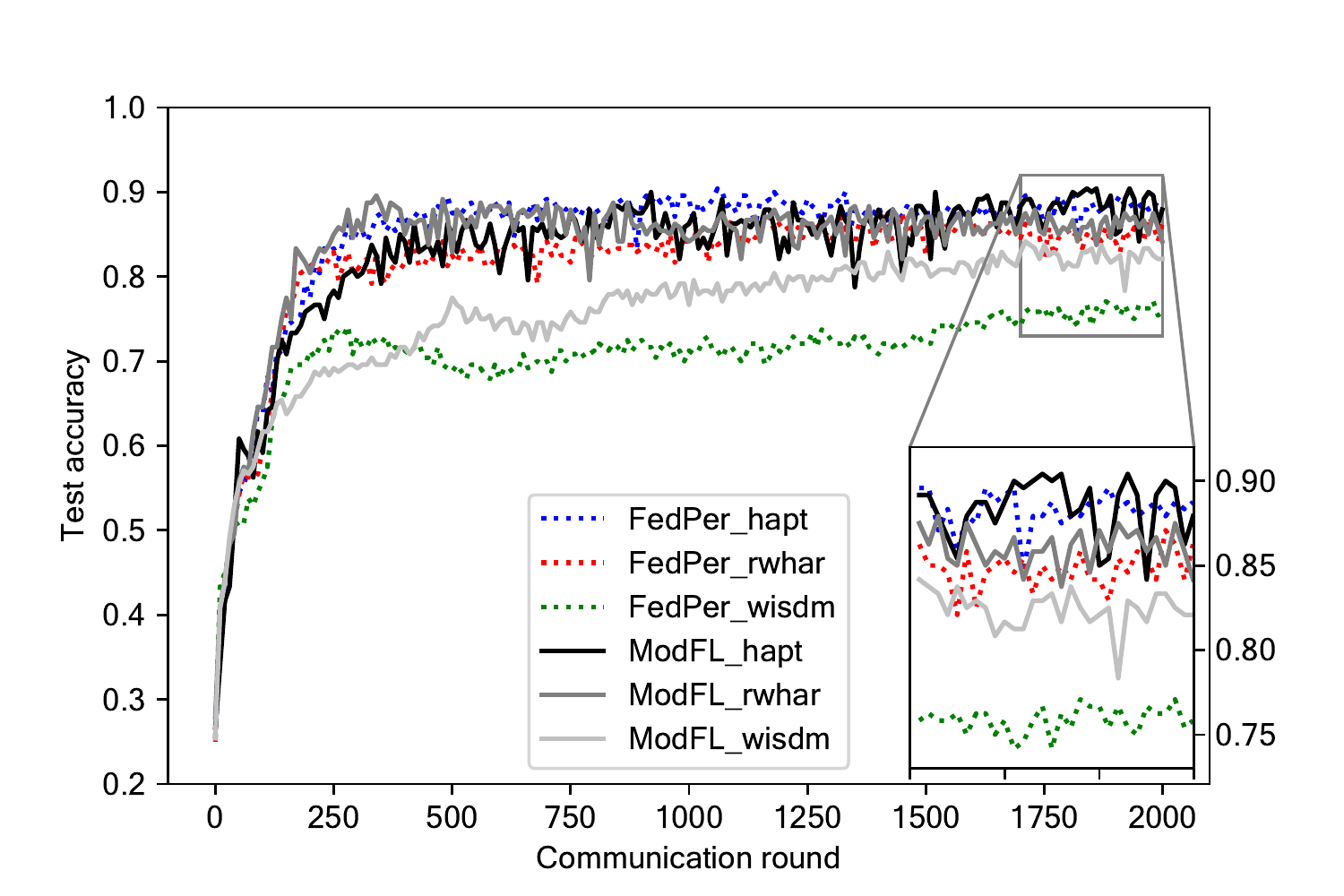}
		 \caption{}
         	 \label{fig:har_results2}
	 \end{subfigure}
	 \hfill
	 \begin{subfigure}{0.32\textwidth}
	         \centering
        	 \includegraphics[width=\textwidth]{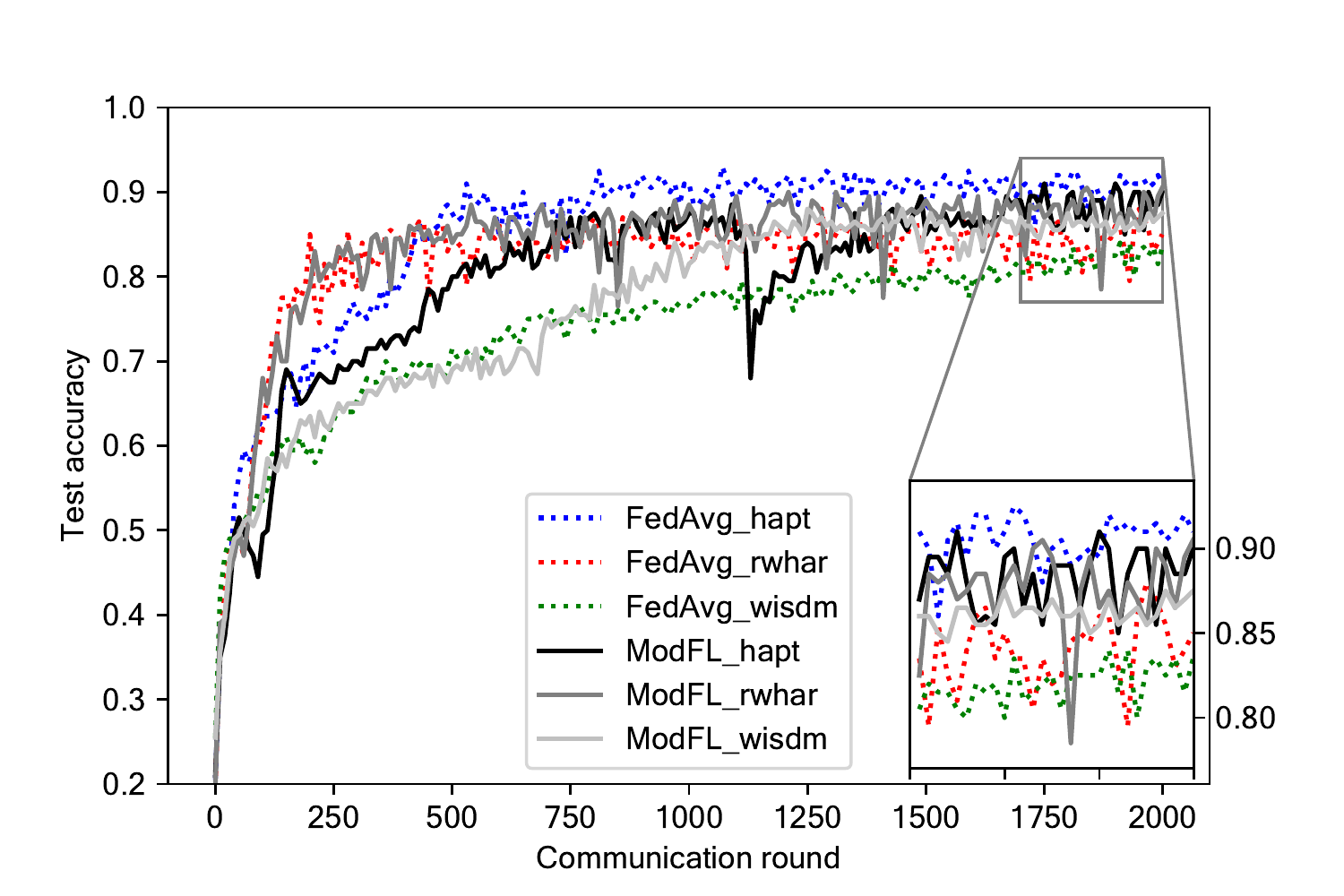}
		 \caption{}
         	 \label{fig:har_results3}
	 \end{subfigure}
     \caption{Performance of \texttt{ModFL} on human activity classification using RNN (the last 300 rounds are zoomed in):
     (a) Non-IID partition with ${P}_{har}=2$ labels compared to \texttt{FedPer}, 
     (b) Non-IID partition with ${P}_{har}=4$ labels compared to \texttt{FedPer}, 
     (c) IID partition compared to \texttt{FedAvg}.
     \label{fig:har_results}
     }
\end{figure*}

\section{Results}
\label{sec:results}
\subsection{Image Classification with CNN}

To study the potential improvement of performance of \texttt{ModFL} over other
federated learning frameworks, we run different number of clients to federate
among. With a finite dataset divided into different clients, a large number of
clients ${N}$ results in less data per client to train the local models. For the
\texttt{FedPer} framework this translates into less data for training the
personalisation layers. In Fig.~\ref{fig:client_vs_acc} (averaged across same dataset clients),
we see that as ${N}$ increases, performance of \texttt{FedPer} decreases. In
contrast, the performance of \texttt{ModFL} suffers marginally as ${N}$ increases.
This is due to the fact that \texttt{ModFL} takes advantage from the federation
of \textit{like-minded} clients belonging to the same operation group.

\begin{table}[t]
	\centering
	\caption{Test accuracy on image classification. Note that FedX is \texttt{FedPer} for non-IID cases and \texttt{FedAvg} for IID case.}
	\label{tab:img_results}
	\begin{tabular}{c | c c c} 
		\multirow{2}{7.5em}{\backslashbox{Models}{${P}_{img}$}} & 3 & 6 & 9 \\ 
		& (non-IID) & (non-IID) & (IID) \\
		\hline
		FedX\_cifar & 71.23\% & 49.75\% & 57.92\% \\ 
		ModFL\_cifar & 78.29\% & 62.82\% & 59.36\% \\
		\hline
		FedX\_stl & 72.09\% & 54.17\% & 62.59\% \\
		ModFL\_stl & 78.46\% & 66.35\% & 63.07\%
	\end{tabular}
\end{table}

For all tested number of clients ${N}$ with ${P}_{img} = \left\{3, 6\right\}$ 
labels, we find that \texttt{ModFL} performs better than \texttt{FedPer} for the CIFAR and
the STL dataset. Fig.~\ref{fig:img_results} shows the test accuracy (averaged across same 
dataset clients) as a function of the number of communication rounds for ${N}=72$ 
clients and different data heterogeneities. In Fig.~\ref{fig:img_results1} the number of
operation groups is $|\{O_j\}|=9$ with ${P}_{img}=3$, here the accuracy of
\texttt{FedPer} for CIFAR and STL starts to stagnate after about 100 communication rounds, 
contrary \texttt{ModFL} continuously improves the accuracy, beyond \texttt{FedPer}, and does 
not seem to stagnate even after 200 rounds. In Fig.~\ref{fig:img_results2} the number of 
operation groups is $|\{O_j\}|=9$ with ${P}_{img}=6$ and again the accuracy 
of \texttt{FedPer} stagnates but at around 50 communication rounds. The \texttt{ModFL} 
accuracy continuously increases even after 200 rounds. Do note that the total number of models 
of \texttt{FedPer} is equal to the number of clients and in the case of \texttt{ModFL} the total 
number of models is 18, \textit{i.e.}, equal to the total number of possible combinations of
configuration modules and operation modules. Fig.~\ref{fig:img_results3} where
${P}_{img} = 9$, we compare \texttt{ModFL} to the \textit{vanilla} \texttt{FedAvg} 
due to this case being IID where \texttt{FedAvg} performs better than \texttt{FedPer}. We 
see that in the IID data case \texttt{ModFL} performs as good as \texttt{FedAvg}, thus the 
proposed protocol implementation deals IID data as good as the \texttt{FedAvg} approach with 
federated learning. The accuracy results are summarised in TABLE~\ref{tab:img_results}. 

\subsection{Human Activity Classification with RNN}
We show results of the \texttt{ModFL} framework applied to time-series data by
evaluating the accuracy of human activity classification using an RNN model.
Fig.~\ref{fig:har_results1} shows the test accuracy (averaged across same dataset 
clients) as a function of the communication rounds for the \texttt{FedPer} and the 
\texttt{ModFL} frameworks evaluated for the HAPT, RWHAR and WISDM datasets 
with a number of different operation groups of $|\{O_j\}|=5$ and level of 
data heterogeneity ${P}_{har}=2$. Here the accuracy between the 
two frameworks is basically the same for the HAPT and RWHAR models. For the 
WISDM model, the accuracies for both frameworks do not overlap as the accuracy 
for both frameworks fluctuates. Fig.~\ref{fig:har_results2} shows the test accuracy 
with a number of different operation groups $|\{O_j\}|=5$ and level of 
data heterogeneity ${P}_{har}=4$. The accuracy curves for both
frameworks overlap for the HAPT and RWHAR datasets and for the WISDM dataset the
curves fluctuate. No clear tenendcy from either of the frameworks can be drawn
from this data. We get similar results for the experiments with all different ${N}$ 
client numbers. Fig.~\ref{fig:har_results3} shows the test accuracy only one operation 
group $|\{O_j\}|=1$ containing all 5 labels, \textit{i.e.}, IID data. We observe 
that for RWHAR, the curves fall into each other, but for the HAPT and the WISDM the 
accuracy curves fluctuate and do not fall onto each other. We attribute the fluctuations 
in the data to sensitivity in initial conditions. The accuracy results are summarised in 
TABLE~\ref{tab:har_results}. 

\begin{table}[t]
	\centering
	\caption{Test accuracy on human activity classification. Note that FedX is \texttt{FedPer} for non-IID cases and \texttt{FedAvg} for IID case.}
	\label{tab:har_results}
	\begin{tabular}{c | c c c} 
		\multirow{2}{7.5em}{\backslashbox{Models}{${P}_{img}$}} & 2 & 4 & 5 \\ 
		& (non-IID) & (non-IID) & (IID) \\
		\hline
		FedX\_hapt & 97.08\% & 88.75\% & 91.00\% \\ 
		ModFL\_hapt & 97.08\% & 87.92\% & 90.00\% \\
		\hline
		FedX\_rwhar & 97.08\% & 86.25\% & 85.00\% \\
		ModFL\_rwhar & 96.25\% & 84.17\% & 90.50\% \\
		\hline
		FedX\_wisdm & 94.17\% & 75.83\% & 83.50\% \\
		ModFL\_wisdm & 92.08\% & 82.08\% & 87.50\% 
	\end{tabular}
\end{table}

Comparing the results in Fig.~\ref{fig:har_results} with the ones in the other
experiments with different amount of ${N}$ clients (not shown), we conclude that
the fluctuations are inherent to the training on the datasets and no conclusion
can be drawn about on which framework is preferably used in with this dataset. What the
data suggests is that \texttt{ModFL} for time-series data is as good as
\texttt{FedPer} for non-IID data and \texttt{FedAvg} for IID data. Moreover,  we
argue that the lack of the performance improvement of \texttt{ModFL} over
\texttt{FedPer} is due to the nature of the dataset; the accuracy of the models
for all three frameworks is relatively high $\approx 0.8$ or higher, even for
the vanilla \texttt{FedAvg}, leaving little room for performance improvement. We
think that to exploit the potential of \texttt{ModFL}, more challenging datasets
than the chosen human activity classification are needed.

\enlargethispage{-0.4in} 

\section{Conclusions}
\label{sec:conclusions}
In this work, we propose \texttt{ModFL} a novel framework to train neural
networks in a federated learning fashion when non-IID data is generated and 
the edge devices are heterogeneous, \textit{i.e.}, they belong to different
generations and thus have different kind of data but perform the same task. We
achieve this by extending the \texttt{FedPer} approach of splitting the neural
networks into configuration modules (layers) and operation modules (layers) and
let the configuration modules federate between same kind of devices while
like-minded users are allowed to federate the operation modules. Results on
CIFAR-10 and STL-10 data indicate that the \texttt{ModFL} protocol for CNN is well
suited to take full advantage of the a federated learning approach among
different groups of peers and show that it outperforms \texttt{FedPer}. It
serves as an alternative to \texttt{FedPer} when the individual clients do not
have sufficient data to train the personalisation layers. The results on the
time-series data using RNN are inconclusive and further research on more
challenging datasets is needed. Further work on more challenging time-series
data is desired to understand better the advantages and limitations of the
proposed framework in this kind of data. 

\section{Acknowledgments}
We thank Anders Vesterberg for insights and useful discussions.  This work was
supported by VINNOVA within the FFI program under contract 2020-02916.


\bibliographystyle{IEEEtran}
\bibliography{IEEEabrv,references} 

\end{document}